\pdfoutput=1

\documentclass[11pt]{article}

\usepackage{acl}

\usepackage{times}
\usepackage{latexsym}

\usepackage[T1]{fontenc}

\usepackage[utf8]{inputenc}

\usepackage{microtype}

\usepackage{graphicx}
\usepackage{float}
\usepackage{array, booktabs,caption, multirow, tabularx}
\usepackage{makecell}
\usepackage{amsfonts} 
\usepackage{amsmath}
\usepackage{subcaption}
\usepackage{courier}

\usepackage{amsthm}
\theoremstyle{definition}
\newtheorem{definition}{Definition}[section]
\DeclareMathOperator*{\argmax}{arg\,max}

\title{BERT is Robust! A Case Against Synonym-Based Adversarial Examples in Text Classification}

\author{Jens Hauser \quad Zhao Meng\thanks{ \; The two authors contributed  equally to this paper.} \quad Damián Pascual\footnotemark[1] \quad \textbf{Roger Wattenhofer} \\
  ETH Zurich, Switzerland \\
    \texttt{\{jehauser, zhmeng, dpascual, wattenhofer\}@ethz.ch} \\
 }

\begin{document}
\maketitle

\begin{abstract}
Deep Neural Networks have taken Natural Language Processing by storm. While this led to incredible improvements across many tasks, it also initiated a new research field, questioning the robustness of these neural networks by attacking them. In this paper, we investigate four word substitution-based attacks on BERT. We combine a human evaluation of individual word substitutions and a probabilistic analysis to show that between 96\% and 99\% of the analyzed attacks do not preserve semantics, indicating that their success is mainly based on feeding poor data to the model.  To further confirm that, we introduce an efficient data augmentation procedure and show that many adversarial examples can be prevented by including data similar to the attacks during training. An additional post-processing step reduces the success rates of state-of-the-art attacks below 5\%. Finally, by looking at more reasonable thresholds on constraints for word substitutions, we conclude that BERT is a lot more robust than research on attacks suggests. 
\end{abstract}

\section{Introduction}

Research in computer vision \citep{szegedy2013intriguing, goodfellow2015explaining} and speech recognition \citep{carlini2018audio} has shown that neural networks are sensitive to changes that are imperceptible to humans. These insights led to extensive research on attacks for creating these so-called adversarial examples, especially in the field of computer vision. Looking for similar issues in NLP is natural, and researchers proposed several different attacks over the last years. However, contrary to computer vision, adversarial examples in NLP are never completely invisible, as discrete characters or words have to be exchanged. This brings up the question: How good are these attacks? Do they reveal issues in current models, or are they just introducing nonsense?

In this paper, we show that despite the general consensus that textual adversarial attacks should preserve semantics, striving for ever-higher success rates seems to be more important when implementing them. We combine a human evaluation with a simple probabilistic analysis to show that between 96\% and 99\% of the adversarial examples on BERT \citep{devlin2019bert} created by four different attacks do not preserve semantics. Additionally, we propose a two-step procedure consisting of data augmentation and post-processing for defending against adversarial examples\footnote{We will release the code with the official publication of this paper.}. While this sounds contradictive at first, the results show that we can eliminate a large portion of the successful attacks by simply including data similar to the adversarial examples and further detect many of the remaining adversarial examples in a post-processing step. Compared to traditional adversarial training strategies, our method is much more efficient and can be used as a baseline defense for researchers looking into new and better attacks.  

\section{Related Work}
\citet{papernot2016crafting} was the first to introduce adversarial examples in the text domain. In the following years, a range of different attacks have been proposed. \citet{alzantot2018generating} use a population-based optimization algorithm for creating adversarial examples, \citet{zhang2019generating} use Metropolis Hastings \citep{metropolis1953equation, hastings1970monte}. Further word substitution based attacks were proposed by \citet{ren2019generating, jin2020bert, li2020bert} and \citet{garg2020bae}. They are discussed in more detail in Section \ref{attacks}.

Regarding adversarial defense, some papers introducing attacks incorporate the created adversarial examples during training \citep{alzantot2018generating, ren2019generating}. However, due to the high cost of running the attacks, they cannot create sufficiently many adversarial examples and achieve only minor improvements in robustness. \citet{wangnatural} suggest Synonym Encoding Method (SEM), a method that uses an encoder that maps clusters of synonyms to the same embedding. Such a method works well but also impedes the expressiveness of the network. \citet{wang2021adversarial} propose a method for fast adversarial training called Fast Gradient Projection Method (FGPM). However, their method is limited to models with non-contextual word vectors as input. On BERT, \citet{meng2021self} use a geometric attack that allows for creating adversarial examples in parallel and therefore leads to faster adversarial training. Another line of work is around certified robustness through Interval Bound Propagation \citep{jia2019certified, huang2019achieving}, but these approaches currently do not scale to large models and datasets.

There is little work criticizing or questioning current synonym-based adversarial attacks in NLP, \citet{morris2020reevaluating} find that adversarial attacks often do not preserve semantics using a human evaluation. They propose to increase thresholds on frequently used metrics for the similarity of word embeddings and sentence embeddings. However, they only investigate a single attack on BERT.

\section{Background}
For a classifier $f: \mathcal{S} \to \mathcal{Y}$ and some correctly classified input $s \in \mathcal{S}$, an adversarial example is an input $s_{pert} \in \mathcal{S}$, such that $sim(s, s_{pert}) \geq t_{sim}$ and $f(s) \neq f(s_{pert})$, where $sim(s, s_{pert}) \geq t_{sim}$ is a constraint on the similarity of $s$ and $s_{pert}$. 
For text classification, $s = \{w^1,w^2,...,w^n\}$ is a sequence of words. Common notions of similarity are the cosine similarity of counter-fitted word vectors \citep{mrkvsic2016counter}, which we will denote as $cos_{cv}(w^i, w_{pert}^i)$ or the cosine similarity of sentence embeddings from the Universal Sentence Encoder (USE) \citep{cer2018universal}, which we will denote as $cos_{use}(s, s_{pert})$. Note that this is a slight abuse of notation since $s$ and $s_{pert}$ are just sequences of words. The notation should be interpreted as follows: We first apply USE to $s$ and $s_{pert}$ to get two sentence vectors and then calculate the cosine similarity. The same holds for $cos_{cv}(w^i, w_{pert}^i)$, where we first get the counter-fitted word vectors of $w^i$ and $w_{pert}^i$. Also, note that whenever we talk about the \textit{cosine similarity of words}, it refers to the cosine similarity of words in the counter-fitted embedding. Similarly, \textit{USE score} refers to the cosine similarity of sentence embeddings from the USE.

\subsection{Attacks}
\label{attacks}
We use four different attacks for our experiments. All of them are based on the idea of exchanging words with other words of similar meaning. The attacks differ in the search method for defining the order of words to replace, in the strategy of choosing the candidate set for replacement words, and in the constraints. To better interpret the results of our analysis, we give a brief summary of the four attacks. Particularly, we are interested in how the attacks build the candidate sets for replacement and in what constraints exist. 
\paragraph{TextFooler}\citet{jin2020bert} propose TextFooler, which builds its candidate set from the 50 nearest neighbors in a vector space of counter-fitted word embeddings. The constraints are $cos_{cv}(w^i, w_{pert}^i) \geq 0.5 \; \forall i$ and $cos_{use}(s, s_{pert}) \geq 0.878$\footnote{The official value is 0.841 on the angular similarity between sentence embeddings, which corresponds to a cosine similarity of 0.878}.
\paragraph{Probability Weighted Word Saliency (PWWS)} PWWS \citep{ren2019generating} uses WordNet\footnote{\url{https://wordnet.princeton.edu/}} synonyms to construct a candidate set. It uses no additional constraints.
\paragraph{BERT-Attack} \citet{li2020bert} suggest an attack based on BERT itself. BERT-Attack uses a BERT masked-language model (MLM) to propose 48 possible replacements. The constraints are $cos_{use}(s, s_{pert}) \geq 0.2$ and a maximum of 40\% of all words can be replaced. 
\paragraph{BAE} \citet{garg2020bae} propose another attack based on a BERT MLM. BAE uses the top 50 candidates of the MLM and tries to enforce semantic similarity by requiring $cos_{use}(s, s_{pert}) \geq 0.936$. 

An attack is successful for a given input $s$, if it finds an adversarial example $s_{pert}$ satisfying all constraints. The \textit{attack success rate} is then defined as the number of successful attacks divided by the number of attempted attacks. 

\section{Setup}
We use the BERT-base-uncased model provided by the Hugging Face Transformers~\citep{wolf2019huggingface} for all our experiments and rely on TextAttack \citep{morris2020textattack} for the implementations of the different attacks.
We fine-tuned BERT for two epochs on AG News and Yelp\footnote{We restricted ourselves to examples in Yelp which have fewer than 80 words} and then randomly sampled 1000 examples from each test-set for running the attacks. The clean accuracies of our models are 94.57\% on AG News and 97.31\% on Yelp. The attack success rates of the different attacks are shown in Table \ref{tab:attack_success_rates}.

It is interesting that BAE, which requires a much higher sentence similarity than BERT-Attack, is a lot less effective despite being otherwise similar. But is a high sentence similarity sufficient to ensure semantic similarity? This is part of what we wanted to investigate using a human evaluation.  

\section{Quality of Adversarial Examples}
To investigate the quality of adversarial examples, we conducted a human evaluation on word substitutions performed by the different attacks. In the following, we call such a word substitution a \textit{perturbation}. A probabilistic analysis is then used to generalize the results on perturbations to attacks. 
\subsection{Human Evaluation}
For the human evaluation, we rely on labor crowd-sourced from Amazon Mechanical Turk\footnote{\url{https://www.mturk.com/}}. We limited our worker pool to workers in the United States and the United Kingdom who completed over 5000 HITs with over 98\% success rate. We collected 100 pairs of [\textit{original word}, \textit{attack word}] for every attack and another 100 pairs for every attack where the context is included with a window size of 11. For the word-pairs, inspired by \citet{morris2020reevaluating}, we asked the workers to react to the following claim: \textit{``In general, replacing the first word with the second word preserves the meaning of the sentence.''} For the words with context, we presented the two text fragments on top of each other, highlighted the changed word, and asked the workers: \textit{``In general, the change preserves the meaning of the text fragment.''} In both cases the workers had seven answers to choose from: ``Strongly Disagree'', ``Disagree'', ``Somewhat Disagree'', ``Neutral'', ``Somewhat Agree'', ``Agree'', ``Strongly Agree''. We convert these answers to a scale from 1-7.

\begin{table}[t]
\small
\centering
\begin{tabularx}{\columnwidth}{@{}ccccc@{}}\toprule
\multirow{2}{*}[-0.1cm]{\textbf{Dataset}} &   \multicolumn{4}{c}{\textbf{Attack Success Rate (\%)}} 
  \\\cmidrule(lr){2-5}
    &  TextFooler & PWWS   & BERT-Attack & BAE   \\ \midrule
  AG News & 84.99 & 64.95 & 79.43 & 14.27 \\
  Yelp & 90.47 & 92.23 & 93.47 & 31.50 \\
\bottomrule
\end{tabularx}
\caption{ \label{tab:attack_success_rates} Attack success rates of the different attacks on fine-tuned BERT-base-uncased models.}
\end{table}
\begin{table*}[t]
\small
\centering 
\begin{tabular}{@{}ccccccc@{}}\toprule
 \multirow{2}{*}[-0.1cm]{\textbf{Attack}} & \multicolumn{3}{c}{\textbf{Word Similarity}} & \multicolumn{3}{c}{\textbf{Text Similarity}}
\\\cmidrule(lr){2-4}\cmidrule(lr){5-7}
           & Avg. (1-7) & Above 5 (\%) & Above 6 (\%)   & Avg. (1-7)  & Above 5 (\%) & Above 6 (\%)\\\midrule
TextFooler    & \textbf{3.88} & \textbf{22} & \textbf{7} & \textbf{3.47} & \textbf{24} & \textbf{12} \\
PWWS        & 3.83 & 21 & 6 & 2.70 & 13 & 6 \\
BERT-Attack & 2.27 & 4 & 4 & 2.55 & 7 & 3\\
BAE        & 1.64 & 0 & 0 & 1.85 & 3 & 2 \\\bottomrule
\end{tabular}
\caption{ \label{tab:human_evaluation} Average human scores on a scale from 1-7 and the percentage of scores above 5 and 6 (corresponding to the answers ``Somewhat Agree'' and ``Agree'') for the different attacks and when the words were shown with (text similarity) or without (word similarity) context.}
\end{table*}

Table \ref{tab:human_evaluation} shows the results of this human analysis. Contrary to what is suggested in papers proposing the attacks, our results show that humans generally tend to disagree that the newly introduced word preserves the meaning. This holds for all attacks and regardless of whether we show the word with or without context. We believe this difference is mainly due to how the text is shown to the judges and what question is posed. For example, asking \textit{``Are these two text documents similar?''} on two long text documents that only differ by a few words is likely to get a higher agreement because the workers will not bother going into the details. Therefore, we believe it is critical to show the passages that are changed.

Regarding the different attacks, it becomes clear from this evaluation that building a candidate set from the first 48 or 50 candidates proposed by a MLM does not work without an additional constraint on the word similarity. The idea of BERT-based attacks is to only propose words that make sense in the context, however, fitting into the context and preserving semantics is not the same thing. The results on BAE further make it clear that a high sentence similarity according to the USE score is no guarantee for semantic similarity. 
PWWS and TextFooler receive similar scores for word similarity, but the drop in score for PWWS when going from word similarity to text similarity indicates that while the synonyms retrieved from WordNet are often somewhat related to the original word, the relation is often the wrong one for the given context. TextFooler receives the highest scores in this analysis, but even for TextFooler, just 22\% and 24\% of the perturbations were rated above 5, which corresponds to ``Somewhat Agree''.
\subsection{Probabilistic Estimation of Valid Attacks}
The human evaluation is based on individual perturbations. An attack usually changes multiple words and therefore consists of multiple perturbations. This begs the question: How many of the successful attacks are actually valid attacks? To answer this question, we need to define valid attacks and valid perturbations. 
\begin{figure}[t]
  \centering
  \includegraphics{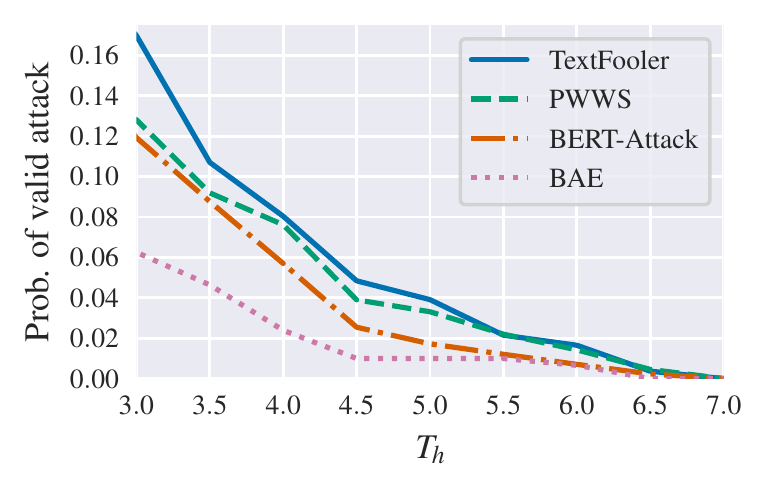}
    \caption{Probability that an attack is valid according to our probabilistic analysis, for the different attacks and for different thresholds $T_h$.}
    \label{fig:p_valid}
\end{figure}
\begin{figure*}[t]
  \centering
  \subfloat{\includegraphics[width=0.45\linewidth]{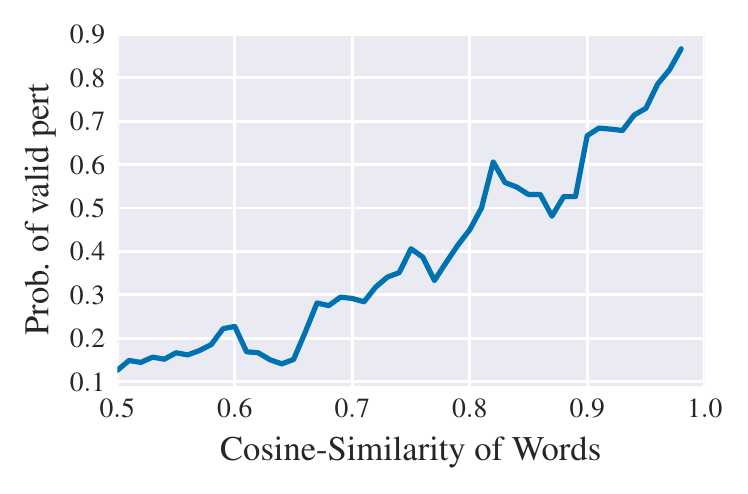}\label{fig:p_valid_vs_cossim}}
  \hfill
  \subfloat{\includegraphics[width=0.45\linewidth]{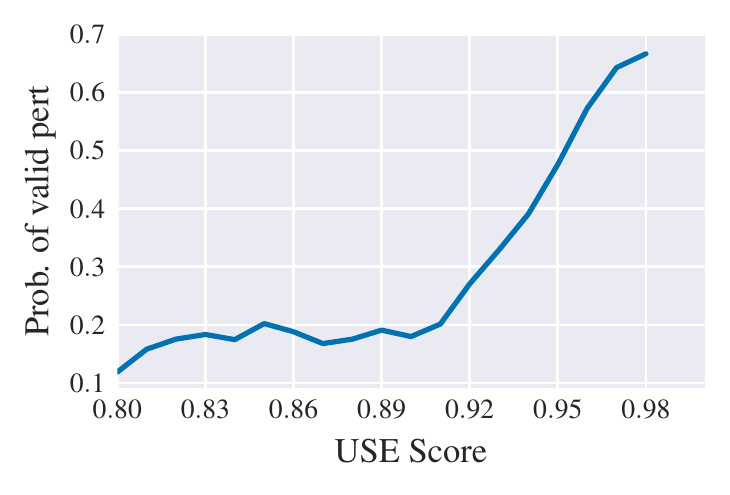}\label{fig:p_valid_vs_use}}
  \caption{The probability that a perturbation is considered valid by a human, as a function of cosine similarity of words (left) and USE score (right). $T_{h}$ is set to 5, i.e. an average score of 5 is required to be considered valid.}
  \label{fig:p_valid_vs_metrics}
\end{figure*}
\theoremstyle{definition}
\begin{definition}[Valid Perturbation]
A \textit{valid perturbation} is a perturbation that receives a human score above some threshold $T_{h}$.
\end{definition}
\begin{definition}[Valid Attack]
A \textit{valid attack} is an attack consisting of valid perturbations only.
\end{definition}
Sensible values for $T_{h}$ are in the range 5-6, which corresponds to ``Somewhat Agree'' to ``Agree''. In order to get an estimate for the percentage of valid attacks, we perform a simple probabilistic analysis. Let $A_{val}$, $P_{val}$ and $A_{val}^i$ denote the events of a valid attack, a valid perturbation and a valid attack consisting of exactly $i$ perturbations. Further, let $p(i)$ denote the probability that an attack perturbs $i$ words. Using that notation, we can approximate the probability that a successful attack is valid as
\begin{equation}
    \begin{split}
    p(A_{val})& = \sum_{i=1}^N p(i)p(A_{val}^i) \\
                & \approx \sum_{i=1}^N p(i)p(P_{val})^i, 
    \end{split}
\end{equation}
where $N$ is the maximal number of allowed perturbations. With the data from Amazon Mechanical Turk and the collected adversarial examples, we can get an unbiased estimate for this probability as 
\begin{equation}
    \hat{p}(A_{val}) = \sum_{i=1}^N \hat{p}(i) \left( \frac{\text{count[$S_{h} \geq T_{h}$]}}{n_{pert}} \right)^i,
\end{equation}

\noindent where $S_{h}$ is the average score of the workers for a perturbation, $n_{pert}$ is the total number of perturbations analyzed by the workers for any given attack, and $\hat{p}(i)$ can be estimated using counts. The results of this analysis are shown in Figure \ref{fig:p_valid} as a function of the threshold $T_{h}$. It can be seen that if we require an average score of 5 for all perturbations, we can expect around 4\% of the successful attacks from TextFooler to be valid, slightly less for PWWS, below 2\% for BERT-Attack, and just around 1\% for BAE. In other words, between 96\% and 99\% of the successful attacks can not be considered valid according to the widely accepted requirement that adversarial examples should preserve semantics.   

This analysis assumes that perturbations are independent of each other, which is not true because every perturbation impacts the following perturbations. Nevertheless, we argue that this approximation tends to result in optimistic estimates on the true number of valid attacks for the following reasons: 
1) When an attack is already almost successful, all attacks except for PWWS try to maximize sentence similarity on the last perturbation, making the last perturbation generally weaker.  2) We strongly assume that in a sentence with multiple changes, a human is generally less likely to say that the meaning is preserved, even if the individual perturbations are considered valid.

\subsection{Metrics vs. Human}
Figure \ref{fig:p_valid_vs_metrics} shows the probability that a perturbation is considered valid (for $T_h = 5$) as a function of cosine similarity of words and as a function of USE score. 
The plots are based on the 400 words with context from the different attacks which were judged by humans. We use left-aligned buckets of size 0.05, i.e., the probability of a valid perturbation for a given cosine similarity $x$ and metric $m \in \{cos_{cv}(\cdot, \cdot), cos_{use}(\cdot, \cdot)\}$, is estimated as
\begin{equation}
      \frac{\text{count[$\left(S_{h} \geq T_{h}\right) \land \left(m \in [x,x+0.05)\right)$]}}{\text{count[$m \in [x,x+0.05)$]}}.
\end{equation}
It can be observed that there is a strong positive correlation between both metrics and the probability that a perturbation is considered valid, confirming both the validity of such metrics and the quality of our human evaluation. However, the exact probabilities have to be interpreted with care, as the analysis based on one variable does not consider the conditional dependence between the two metrics. 

\section{Adversarial Defense}
We have shown that current attacks use lenient constraints and, therefore, mostly produce adversarial examples that should not be considered valid, but finding suitable thresholds on the constraints is difficult. \citet{morris2020reevaluating} try to find these thresholds by choosing the value where humans ``Agree'' (on a slightly different scale) on average and find thresholds of 0.90 on the word similarity and 0.98 on the sentence similarity score. However, this misses all the perturbations which were considered valid by the workers at lower scores (see Figure \ref{fig:p_valid_vs_metrics}). Before discussing other thresholds, we show that we can avoid many adversarial examples even for low thresholds. 

Our procedure consists of two steps, where the first step prepares for the second. The first step is a data augmentation procedure and looks as follows: 
\paragraph{Step 1}
\begin{enumerate}
    \item[a)] Initialize thresholds $t_{rr} \in \left(0,100\right]$ for the maximal percentage of words to augment, and $t_{cv} \in \left(0,1\right)$ for a threshold on cosine similarity of words.
    \item[b)]  During training of the model, for every batch, calculate the gradients to get the $t_{rr}$ percent of most important words for every input. The union of the words considered as stop-words by the four attacks is filtered out.
    \item[c)]  Then, for every word marked as important according to b), a candidate set $\mathcal{C}$ is built out of all words in a counter-fitted embedding with cosine similarity greater than $t_{cv}$.
    \item[d)]  To account for the fact that all attacks tend to favor words with low cosine similarity (see Appendix \ref{sec:cos_sim}), the replacement $v_i \in \mathcal{C}$ for the original word $w$ is chosen with probability:
    \begin{equation}
    p(v_i) = \frac{1-\textit{cos}_{cv}(w, v_i)}{\sum_{v_j \in \mathcal{C}} 1-\textit{cos}_{cv}(w, v_j)}.
    \end{equation}
    This skews the probability towards words with lower cosine similarity.
    \item[e)] Finally, the perturbed batch with the changed words is concatenated to the original batch
\end{enumerate}
\begin{table*}[t]
\small
\centering
\begin{tabular}{@{}ccccccc@{}}\toprule
\multirow{2}{*}[-0.1cm]{\textbf{Dataset}} & \multirow{2}{*}[-0.1cm]{\textbf{Method}} & \multirow{2}{*}[-0.1cm]{\makecell{\textbf{Clean} \\ \textbf{Acc. (\%)}}} & \multicolumn{4}{c}{\textbf{Attack Success Rate (\%)}} 
  \\\cmidrule(lr){4-7}
    & &   & TextFooler & PWWS$_{cv50}$  & BERT-Attack$_{cv50}$ & BAE$_{cv50}$
    \\ \midrule
  \multirow{5}{*}{AG News} & Normal & 94.57 & 84.99 & 16.38 & 20.72 & 0.32 \\
   & DA & \textbf{94.82}       & 52.37 & 10.73 & 18.61 & -- \\
    & DA+PP & 93.84 $\pm$ 0.07  & \textbf{3.93} $\pm$ 0.41 & \textbf{2.55} $\pm$ 0.31 & \textbf{3.73} $\pm$ 0.29 & -- \\
    & DA+MA$_5$ & 93.72 $\pm$ 0.12  & 14.11 $\pm$ 0.48 & 4.61 $\pm$ 0.41 & 7.52 $\pm$ 0.48 & -- \\
    & Normal+PP & 87.89 $\pm$ 0.16 & 10.32 $\pm$ 0.48 & 5.0 $\pm$ 0.31 & 5.59 $\pm$ 0.36 & -- \\
    \midrule
  \multirow{5}{*}{Yelp} & Normal & \textbf{97.31} & 90.47 & 33.26 & 49.53 & 0.41\\
   & DA & 97.10         & 29.79 & 10.52 & 16.49 & -- \\
    & DA+PP & 96.59 $\pm$ 0.06  & \textbf{4.37} $\pm$ 0.39 & \textbf{2.54} $\pm$ 0.15 & \textbf{4.86} $\pm$ 0.33 & -- \\
    & DA+MA$_5$ & 95.40 $\pm$ 0.10  & 10.23 $\pm$ 0.59 & 4.62 $\pm$ 0.36 & 7.38 $\pm$ 0.38 & -- \\
    & Normal+PP & 94.50 $\pm$ 0.08  & 6.07 $\pm$ 0.47 & 5.22 $\pm$ 0.48 & 7.35 $\pm$ 0.61 & -- \\

\bottomrule
\end{tabular}
\caption{ \label{tab:defense_procedure_nversions_8} Effectiveness of defense procedure for different attacks modified with constraint on cosine-similarity of words.}
\end{table*}
The data augmentation procedure makes the model more robust against attack words with cosine similarity greater $t_{cv}$. If we expect BERT to be robust against these kinds of replacements, this is the least we should do. Otherwise, we cannot expect the model to generalize to the attack's input space, which is significantly larger than the input space during fine-tuning. 

We can further improve the robustness with a post-processing step that builds on this robustness to random substitutions.
\paragraph{Step 2}
\begin{enumerate}
    \item[a)] For every text that should be classified, $N$ versions are created where $t_{rr} \%$ of the words (which are not stop-words) are selected uniformly at random and are exchanged by another uniformly sampled word from a candidate set $\mathcal{C}$ consisting of all words with cosine-similarity above $t_{cv}$. 
    \item[b)] The outputs of the model (logits) are added up for the $N$ versions and the final prediction is made according to the maximum value. Formally, let $l_j(s)$ denote the value of the $j$-th logit for some input $s$. Then the prediction $y_{pred}$ is made according to
    \begin{equation}
        \label{eq:y_pred}
        y_{pred} = \argmax_j \sum_{i=1}^N l_j(s_i).
    \end{equation}
\end{enumerate}
This procedure can be applied for any threshold $t_{cv} \in (0,1)$, but it only makes sense if we expect an attack to use the same or a higher threshold. We always set $t_{cv}$ to the same value as the attack uses. Further, we set $t_{rr}=40$ and $N=8$ in all our experiments, and we use the same thresholds for both steps.
\section{Defense Results}
In Table \ref{tab:defense_procedure_nversions_8}, we show the effect of the procedure on the different attacks modified with the constraint that the cosine-similarity between original word and attack word should be above 0.5. The notation is the following: Normal stands for a model fine-tuned normally. DA stands for a model fine-tuned with data augmentation, and PP stands for post-processing. MA$_{5}$ is a baseline for our post-processing procedure that replaces 5\% of all tokens with the \texttt{[MASK]} token (see Appendix \ref{sec:mask_baseline}). The results show that up to two-thirds of the attacks can be prevented using data augmentation. This indicates that adversarial examples for text classification are closely related to the data on which the model is fine-tuned. The attacks try to create examples that are out-of-distribution with respect to the training data. Additionally, between 70\% and 92\% of the attacks can be reverted using our post-processing procedure, resulting in attack success rates below 5\% for all attacks. For TextFooler, this corresponds to a decrease in attack success rate of more than 95\%. Because the post-processing step is probabilistic, we ran it ten times for every combination of dataset and attack. We show the mean and standard deviation of the ten resulting attack success rates. Compared to the mask-baseline, our post-processing procedure can revert significantly more attacks while having a smaller impact on the clean accuracy. Table \ref{tab:defense_procedure_nversions_8} also shows that the post-processing step should always be preceded by data augmentation. While applying post-processing in isolation still reverts many attacks, the clean accuracy drops significantly, especially on AG News. 

\begin{table*}[t]
\small
\centering
\begin{tabular}{@{}ccccccc@{}}\toprule
\multirow{2}{*}[-0.1cm]{\textbf{Dataset}} & \multirow{2}{*}[-0.1cm]{\textbf{Method}} &  \multicolumn{5}{c}{\textbf{Attack Success Rate (\%)}} 
  \\\cmidrule(lr){3-7}
    &  & TF$_{cv50}$   & TF$_{cv50}^{use88}$ & TF$_{cv70}^{use85}$  & TF$_{cv70}^{use90}$ & TF$_{cv80}^{use90}$   \\ \midrule
  \multirow{3}{*}{AG News} & Normal & 88.79 & 24.95 & 22.52 & 11.63 & 7.51 \\
   & DA  & 55.58 &  16.11 & 10.79 & 7.12 & 4.50\\
    & DA+PP & 4.49 $\pm$ 0.39  & 3.31 $\pm$ 0.28 & 2.07 $\pm$ 0.16  & 1.91 $\pm$ 0.17 & 0.99 $\pm$ 0.17\\
    \midrule
  \multirow{3}{*}{Yelp} & Normal & 91.40  & 49.22 & 42.59  & 25.18 & 11.09 \\
   & DA      & 38.46   & 13.74  & 10.34 &  7.78 & 2.87\\
    & DA+PP  & 5.04 $\pm$ 0.35 & 3.9 $\pm$ 0.34 & 2.12 $\pm$ 0.21 & 2.28 $\pm$ 0.17 & 0.71 $\pm$ 0.13\\
\bottomrule
\end{tabular}
\caption{ \label{tab:defense_procedure_at} Effectiveness of defense procedure for different combinations of thresholds.}
\end{table*}

\subsection{Adjusted Thresholds}
Table \ref{tab:defense_procedure_nversions_8} shows that with the constraint on cosine similarity of words added, TextFooler is by far the most effective attack, at least before post-processing. There is a simple reason for this, TextFooler already has that constraint and is the only attack out of the four to choose its candidate set directly from the counter-fitted embedding used to calculate the cosine similarity. On the other end of the spectrum, BAE's attacks success rate drops close to zero. This is because the intersection of the set of words proposed by the MLM, the set of words with cosine similarity greater than 0.5, and the set of words keeping the USE score above 0.936 is small and leaves the attack not much room. A similar observation can be made for PWWS, although not as pronounced. 

However, there is one more reason why TextFooler is more effective compared to the other attacks, despite an additional constraint on the USE score. While attacking a piece of text, this constraint on the USE score is not checked between the current perturbed text $s_{pert}$ and the original text $s$, but instead between the current perturbed text $s_{pert}$ and the previous version $s_{pert}^\prime$. This means that by perturbing one word at a time, the effective USE score between $s$ and $s_{pert}$ can be a lot lower than the threshold suggests. When discussing the effect of raising thresholds to higher levels, we do so by relying on TextFooler as the underlying attack because it is the most effective, but we adjust the constraint on the USE score to always compare to the original text. We believe this is the right way to implement this constraint, and more importantly, it is consistent with how we gathered data from Amazon Mechanical Turk. 

Table \ref{tab:defense_procedure_at} shows the results from our defense procedure when the thresholds on TextFooler are adjusted. TF$_{cv50}$ corresponds to TextFooler without the constraint on the USE score. Comparing with Table \ref{tab:defense_procedure_nversions_8} confirms that the original implementation of the USE constraint only had a small impact. TF$_{cv50}^{use88}$ corresponds to TextFooler with $cos_{cv}(w^i, w_{pert}^i) \geq 0.5 \; \forall i$ and $cos_{use}(s, s_{pert}) \geq 0.88 \; (0.878 \text{ to be precise})$, the same thresholds as in the original implementation, but without allowing to drift away from the original text as discussed above. This already decreases the attack success rate significantly. Using data augmentation, we can decrease the attack success rate by more than a factor of 5 compared to what we saw originally (84.99 to 16.11 and 90.47 to 13.74). This shows that by preventing TextFooler from using that little trick and some data augmentation, we can decrease the attack success rate to values far from the ones suggested in their paper. When increasing the thresholds on the constraints (compare to Figure \ref{fig:p_valid_vs_metrics} to see that these are still not particularly strong constraints), it becomes even more evident that BERT is a lot more robust than work on attacks suggests. Especially if we allow for post-processing.
\begin{table*}[t]
\small
\centering
\begin{tabular}{@{}cccccccc@{}}\toprule
\multirow{2}{*}[-0.1cm]{\textbf{Dataset}} & \multirow{2}{*}[-0.1cm]{\textbf{Method}} & \multirow{2}{*}[-0.1cm]{\makecell{\textbf{Clean} \\ \textbf{Acc. (\%)}}} &
\multirow{2}{*}[-0.1cm]{\makecell{\textbf{Training} \\ \textbf{Time (h:min)}}} &
\multirow{2}{*}[-0.1cm]{\textbf{Epochs}} &
\multicolumn{3}{c}{\textbf{Attack Success Rate (\%)}} 
  \\\cmidrule(lr){6-8}
    & & & &  & TextFooler & PWWS$_{cv50}$  & BERT-Attack$_{cv50}$
    \\ \midrule
  \multirow{4}{*}{AG News} & Normal & 94.57 & 0:19 & 2 & 84.99 & 16.38 & 20.72 \\
   & DA & \textbf{94.82}  & 5:33  & 12   & 52.37 & 10.73 & 18.61  \\
    & ADV & 92.83 & 160:15 & 12 & \textbf{34.54} & \textbf{6.50} & \textbf{9.38} \\
    & ADV$_{naive}$ & 94.26 & 45:14 & 2 & 56.20 &  12.50 & 17.44  \\
    \midrule
  \multirow{4}{*}{Yelp} & Normal & \textbf{97.31} & 0:32 & 2 & 90.47 & 33.26 & 49.53 \\
   & DA & 97.10  & 9:08   & 12    & \textbf{29.79} & \textbf{10.52} & \textbf{16.49}  \\
    & ADV & 95.94  & 107:56 & 5 & 59.52 & 14.64 & 25.52 \\
    & ADV$_{naive}$ & 96.65 & 56:53 & 2 & 95.12 & 33.09 & 47.61 \\

\bottomrule
\end{tabular}
\caption{ \label{tab:da_vs_adv} Comparison of data augmentation and adversarial training.}
\end{table*}
\subsection{Comparing data augmentation with adversarial training}
While adversarial training provides the model with data from the true distribution generated by an attack, our data augmentation procedure only approximates that distribution. The goal is to trade robustness for speed. However, it turns out that our procedure can even be superior to true adversarial training in some cases. We compare to two different strategies for adversarial training. ADV$_{naive}$ denotes the simplest procedure for adversarial training in text classification: collect adversarial examples on the training set and then train a new model on the extended dataset consisting of both adversarial examples and original training data. We used TextFooler to collect these adversarial examples. On the complete training set, this resulted in 103'026 adversarial examples on AG News and 179'335 adversarial examples on Yelp. For a more sophisticated version for adversarial training, we follow \citet{meng2021self} by creating adversarial examples on-the-fly during training. We denote this method as ADV (corresponds to ADV in their paper).

A comparison of the results on AG News and Yelp is shown in Table \ref{tab:da_vs_adv}. Interestingly, ADV$_{naive}$ did not result in an improvement on Yelp. We hypothesize that this is because Yelp is easier to attack, resulting in weaker training data for the extended dataset. For example, 26\% of the created adversarial examples on Yelp differ by only one or two words from the original text, on AG News this holds for just 11\% of the adversarial examples. Furthermore, the average word replace rate on Yelp is 16\% compared to 24\% on AG News. The same argument would also explain why, quite surprisingly, we reach higher robustness on Yelp with our data augmentation procedure compared to ADV. To be fair, it must be said that we did not train ADV until convergence on Yelp due to computational constraints. Overall, lower computation time is precisely the biggest advantage of our method. Considering that the training data increases by a factor of two, the overhead per epoch is only around 50\% compared to normal training.  

\section{Limitations}
In practice, the post-processing step cannot be decoupled from a black-box attack. It would be interesting to see how successful an attack is when the whole system, including post-processing, is regarded as a single black-box model. We hypothesize that it would remain challenging because the attacker can rely much less on its search method for finding the right words to replace.

The method is also not applicable if a deterministic answer is required. However, in many applications such as spam filters or fake news detection, we are only interested in making a correct decision as often as possible while being robust to a potential attack. 
\section{Discussion \& Conclusion}
Using a human evaluation, we have shown that most perturbations introduced through adversarial attacks do not preserve semantics. This is contrary to what is generally claimed in papers introducing these attacks. We believe the main reason for this discrepancy is that researchers working on attacks have not paid enough attention to preserving semantics because attacks with new state-of-the-art success rates are easier to publish.  However, in order to find meaningful adversarial examples that could help us better understand current models, we need to get away from that line of thinking. For example, 10-20\% attack success rate with valid adversarial examples and a good analysis on them is much more valuable than 80-90\% attack success rate by introducing nonsensical words. We hope this work encourages researchers to think more carefully about appropriate perturbations to text which do not change semantics. 

Our results on data augmentation show that a significant amount of adversarial examples can be prevented when including perturbations during training that could stem from an attack. It is debatable whether changing 40\% of the words with a randomly chosen word from a candidate set still constitutes a valid input, but this is only necessary because the attacks have that amount of freedom. The more appropriate the allowed perturbations for an attack, the more appropriate is our data augmentation procedure, which can easily be adapted for other candidate sets. Compared to adversarial training, our method scales to large datasets and multiple epochs of training, making it an excellent baseline defense method for researchers working on new attacks and defenses. The post-processing step completes our defense procedure and shows that attacks can largely be prevented in a probabilistic setting without a severe impact on the clean accuracy.  In practice, this means that most attacks can at least be detected. Whether or not this two-step procedure will prevent the same amount of attacks when the whole model is considered a probabilistic black-box is up for future investigation.

\bibliography{anthology,custom}
\bibliographystyle{acl_natbib}

\clearpage
\appendix

\section{Number of versions in post-processing}
In order to understand the impact of the number of versions $N$ created during the post-processing step, we can make the following analysis: 
Let us consider the augmented inputs as instances of a discrete random variable $X$. For $x \in X$ and a classification problem with $K$ classes, let $l_{correct}(x)$ denote the value of the logit corresponding to the correct label and $l_{j}(x)$ denote the value of the $j$-th logit corresponding to a wrong label, such that $j \in \{1,...,K-1\}$. We are only interested in the differences $g_j(x) = l_{correct}(x) - l_{j}(x)$. Ideally, we would like to make a decision based on the expectations of $g_j(X)$. An attack should be reverted if and only if
\begin{equation}
    \label{eq:post_exp}
    \mathrm{E}[g_j(X)] = \sum_{x \in X} g_j(x) p_X(x) \geq 0 \quad \forall j,
\end{equation}
where $p_X(x) = \frac{1}{|X|}$. 
Because we cannot enumerate over all instances $x$, we approximate this with sums over just $N$ instances 
\begin{equation}
    \label{eq:post_est_expt}
    \sum_{i=1}^N \frac{g_j(x_i)}{N} \geq 0 \quad \forall j.
\end{equation}
\begin{table}[t]
\small
\centering
\begin{tabularx}{\columnwidth}{@{}ccccc@{}}\toprule
\multirow{2}{*}[-0.1cm]{\textbf{Dataset}} & 
\multirow{2}{*}[-0.1cm]{\textbf{N}} & 

\multicolumn{3}{c}{\textbf{Reverted Attacks (Mean/Std) (\%)}}
\\\cmidrule(lr){3-5}
         &  & TextFooler & PWWS$_{cv50}$   & BERT-Att$_{cv50}$ \\ \midrule
\multirow{4}{*}{\makecell{AG \\ News}} & 4   & 92.13 / 0.65 & 75.39 / 3.35 & 78.7 / 1.94  \\
& 8  & 92.49 / 0.79 & 76.27 / 2.87 & 79.94 / 1.54 \\
& 16 & 92.81 / 0.53 & 78.24 / 1.95 & 80.17 / 0.85 \\
& 32   & 92.97 / 0.24 & 76.57 / 1.61 & 81.07 / 0.88 \\
\midrule
\multirow{4}{*}{Yelp} & 4     & 83.94 / 1.49 & 74.31 / 3.28 & 68.56 / 3.02  \\
& 8     & 85.33 / 1.32 & 75.88 / 1.4 & 70.5 / 1.97 \\
& 16     & 85.81 / 1.26 & 76.37 / 1.88 & 70.81 / 1.12 \\
& 32   & 86.26 / 0.74 & 76.96 / 0.79 & 71.31 / 2.16 \\

\bottomrule
\end{tabularx}
\caption{ \label{tab:defense_procedure_versions} Effectiveness of post-processing for different number of versions.}
\end{table}
These are unbiased estimates of the expectations in \eqref{eq:post_exp} for any choice of $N$. By multiplying with $N$ and plugging in the definition of $g_j(x)$, it can be verified that a decision based on \eqref{eq:post_est_expt} reverts the same attacks as a decision based on \eqref{eq:y_pred}. The expectation estimates become more and more accurate as we increase $N$. Since we are making a discrete decision based on whether the expectations are $\geq$ 0, the estimate is more likely to be correct with more samples. If we assume that the true expectation is positive in most cases, this means we can generally expect a higher number of reverted attacks for higher $N$. Being more precise on the estimate also means we generally tend to make the same decision every time on the same example, therefore reducing the variance in the reverted attack rate. Table \ref{tab:defense_procedure_versions} shows results on reverted attacks for 4, 8, 16 and 32 versions and generally confirms this. However, the results are already quite good with just four versions, so this is a trade-off between speed and accuracy, as creating $N$ versions increases the batch size during inference by a factor $N$. 
\section{Baseline for post-processing}
\label{sec:mask_baseline}
Instead of replacing words with other words in Step 2 of our defense procedure, one could also think of other ways of slightly perturbing the adversarial examples to flip the label back to the correct one. To show that our method is superior to such simple perturbations, Table \ref{tab:post_processing_mask} shows the results of a baseline procedure in which we replace randomly chosen words with the \texttt{[MASK]} token. Indeed, averaged over TextFooler, PWWS, and BERT-Attack, up to 63\% of the adversarial examples on AG News can be reverted by masking just 5\% of the words. However, further improving on that by masking more tokens fails, and the clean accuracy drops substantially. This is contrary to our procedure, in which we exchange 40\% of the words with just a minimal decrease in accuracy.

\begin{table}[t]
\small
\centering
\begin{tabularx}{\columnwidth}{@{}cccc@{}}\toprule

\textbf{Dataset} & \textbf{Method} & \textbf{Clean Acc. (\%)} & \textbf{Reverted (\%)}
    \\ \midrule
  \multirow{4}{*}{AG News} & MA$_5$ & 93.62 & 63.24   \\
   & MA$_{10}$ & 92.14 &  62.76  \\
    & MA$_{20}$ & 87.30  & 57.34\\
    & MA$_{30}$ & 76.25 &  50.01\\ 
    \midrule
  \multirow{4}{*}{Yelp} & MA$_5$ & 95.19  & 59.0\\
   & MA$_{10}$ & 93.98 & 61.42  \\
    & MA$_{20}$ & 90.53 &  60.83 \\
    & MA$_{30}$ & 86.91 &  59.25 \\
\bottomrule
\end{tabularx}
\caption{ \label{tab:post_processing_mask} By masking random tokens instead of exchanging words, more than half of the attacks can be reverted. However, the clean accuracy drops.}
\end{table}

\section{Word Frequencies}
We observe that attacks frequently introduce words that rarely occur during training. Table \ref{tab:median_word_frequencies} shows median word occurrences (Occ. column) of original words and attack words in the training set for different attacks. The results are quite striking and a further justification for using data augmentation. It is also interesting to see that BERT-Attack acts differently in that regard. We assume this is because BERT-Attack has the weakest constraints (no constraint on cosine similarity of words and a weak constraint on USE). This could allow BERT-Attack to find more effective perturbations than other attacks that have to choose from a set of more similar words and then rely on the ones the model does not know. 

Table \ref{tab:median_word_frequencies} further shows that attacks often use words with higher relative frequency in other classes. Column GT reveals the percentage of times that the original words and attack words have the highest relative frequency (word occurrences in class divided by the total number of words in the same class) in the ground truth class. It can be observed that attacks often introduce words with higher relative frequency in a different class. This is an interesting observation as no one would be surprised by the success of such perturbations if we were dealing with a bag-of-words model.
\section{Cosine Similarities of Words}
\label{sec:cos_sim}
\begin{figure}[t]
\begin{subfigure}{.99\textwidth}
  \includegraphics[scale=0.49]{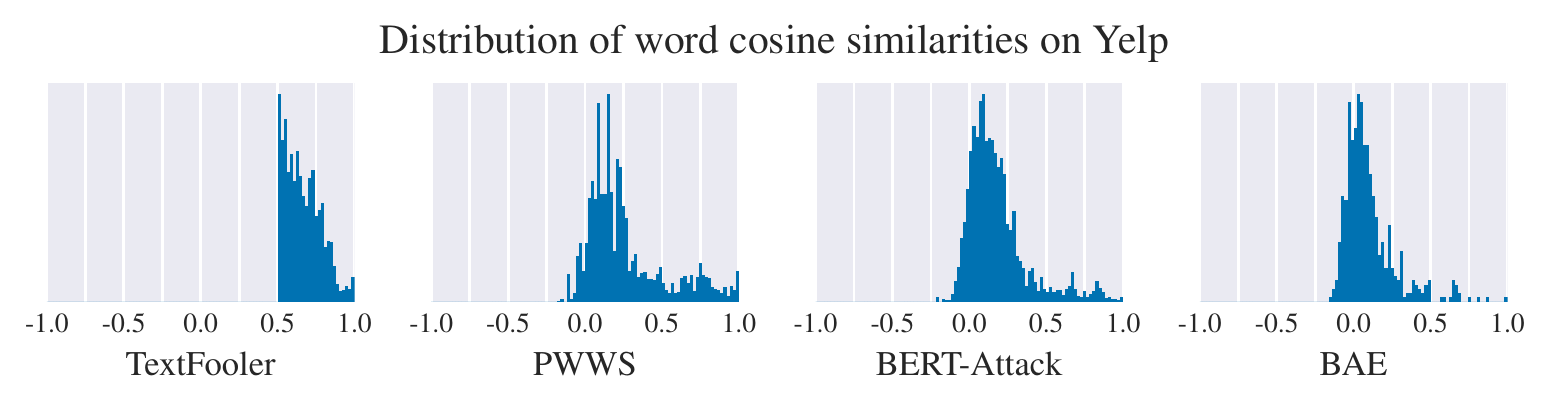}
\end{subfigure}
\begin{subfigure}{.99\textwidth}
  \includegraphics[scale=0.49]{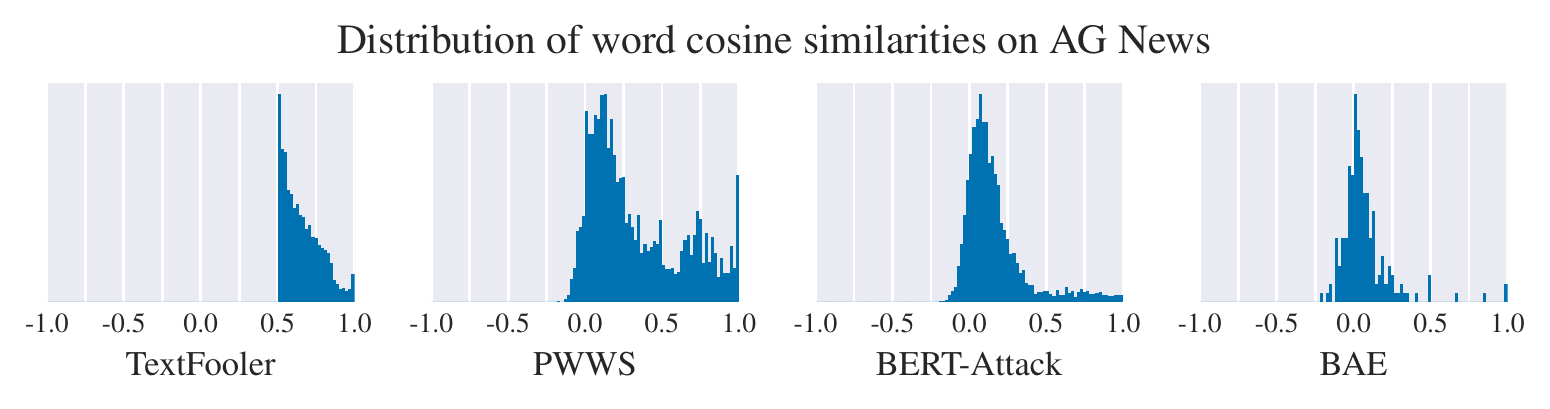}
\end{subfigure}
\caption{\label{fig:cosine_similarities} Distribution of cosine similarities of words.}
\end{figure}

In a counter-fitted embedding, perfect synonyms are supposed to have a cosine similarity of 1 and perfect antonyms are supposed to have a cosine similarity of 0. Figure \ref{fig:cosine_similarities} shows the distribution of cosine similarities for the four attacks on both datasets. 

\begin{table}[t]
\small
\centering
\begin{tabularx}{\columnwidth}{@{}cccccc}
\toprule
\multirow{2}{*}[-0.1cm]{\textbf{Dataset}} & \multirow{2}{*}[-0.1cm]{\textbf{Attack}} & \multicolumn{2}{c}{\textbf{Orig. Word}} & \multicolumn{2}{c}{\textbf{Att. Word}}
\\ \cmidrule(lr){3-4} \cmidrule(lr){5-6}
    &    & Occ. & GT (\%)   & Occ. & GT (\%)   \\ \midrule
\multirow{4}{*}{\makecell{AG \\ News}} & TextFooler   & 736 & 67.31 & 18 & 24.63  \\
                         & PWWS   & 889 & 60.04 & 24 & 16.06  \\
                         & BERT-Att.   & 585 & 65.92 & 344 & 22.91  \\
                         & BAE   & 617 & 52.66 & 4 & 9.31  \\
\midrule
\multirow{4}{*}{Yelp} & TextFooler   & 4240 & 72.79 & 19 & 44.60  \\
                      & PWWS   & 5715 & 74.56 & 13 & 33.76  \\
                      & BERT-Att.   & 4521 & 75.27 & 3398 & 35.55  \\
                      & BAE   & 4601 & 76.03 & 44 & 41.87  \\

\bottomrule
\end{tabularx}
\caption{ \label{tab:median_word_frequencies} Median word occurrences of original words and attack words in training set (Occ.) and percentage of times that words have the highest relative frequency in ground truth class (GT).}
\end{table}

\section{Details on Human Evaluation}
\begin{figure}[t!]
  \centering
  \includegraphics[width=0.48\textwidth]{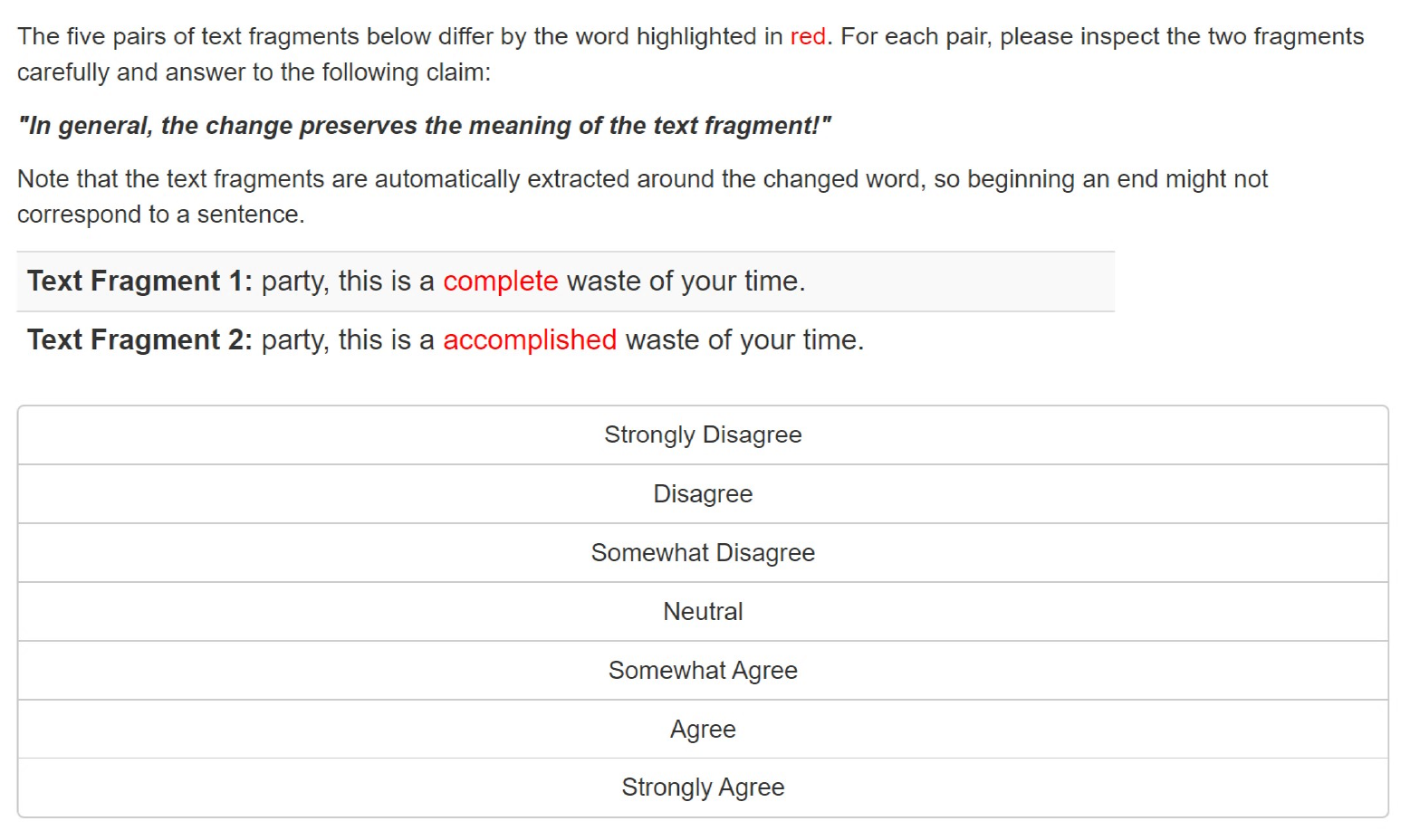}
  \caption{Screenshot of the human evaluation used to evaluate words with context.}
  \label{fig:screenshot_human_analysis_word_context}
\end{figure}

\begin{figure}[t!]
  \centering
  \includegraphics[width=0.48\textwidth]{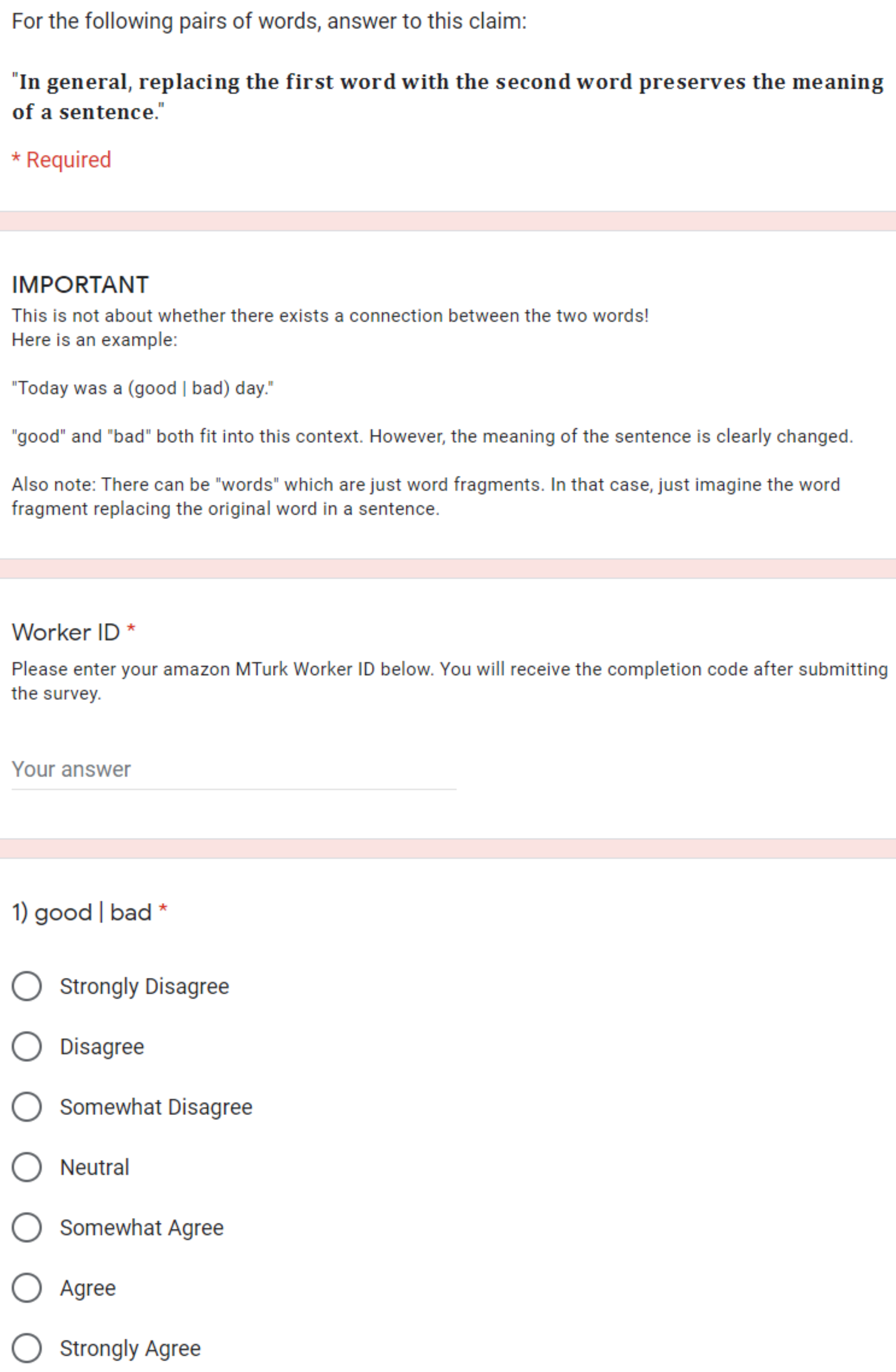}
  \caption{Screenshot of the Google form used to evaluate similarity of words.}
  \label{fig:screenshot_human_analysis_word_pairs}
\end{figure}

We relied on workers with at least 5000 HITs and over 98\% success rate. For the word-pairs, we showed the workers 100 pairs of words in a google form. In order to ensure a good quality of work, we included some hand-designed test cases at several places and rejected workers with strange answers on these word-pairs. These test cases were [\textit{good}, \textit{bad}], [\textit{help}, \textit{hindrance}] (expected answer ``Strongly Disagree'' or ``Disagree'') and [\textit{sofa}, \textit{couch}], [\textit{seldom}, \textit{rarely}] (expected answer ``Strongly Agree'' or ``Agree''). In a first test run, surprisingly, many workers agreed on antonyms like good and bad, which is why we added a note with an example and emphasized that this is about whether the meaning is preserved and not about whether both words fit into the same context. Workers were paid 2.0\$ for one HIT with 100 pairs and 4 test cases. We showed every pair of words to ten workers and calculated the mean. A screenshot of the form can be found in Figure \ref{fig:screenshot_human_analysis_word_pairs}. 
For the words with context, we used the amazon internal form because it allowed for a clearer presentation of the two text fragments (see Figure \ref{fig:screenshot_human_analysis_word_context}). We always presented five pairs of text fragments in one HIT and rejected workers that submitted the hit within less than 60s to ensure quality. Workers were paid 0.5\$ for one HIT with five pairs. We showed every pair of text fragments to five workers and calculated the mean.

\section{Datasets}
For our experiments, we use two different text classification datasets: AG News and Yelp. On Yelp, we only used the examples consisting of 80 words or less. Especially comparing to ADV would have been much harder otherwise. Statistics of the two datasets are displayed in Table \ref{tab:datasets}.
\begin{table}[H]
\small
\centering
\begin{tabularx}{\columnwidth}{ccccc}\toprule
          \textbf{Dataset} & \textbf{Labels} & \textbf{Train}  & \textbf{Test}  & \textbf{Avg Len}   \\ \midrule
AG News   & 4 & 120'000 & 7'600 & 43.93 \\
Yelp   & 2 & 199'237 & 13'548 & 45.69 \\

\bottomrule
\end{tabularx}
\caption{ \label{tab:datasets} Statistics of the two datasets.}
\end{table}
\noindent\textbf{AG News} \citep{zhang2015character} is a topic classification dataset. It is contructed out of titles and headers from news articles categorized into the four classes World, Sports, Business, and Sci/Tech.

\noindent\textbf{Yelp} \citep{zhang2015character} is a binary sentiment classification dataset. It contains reviews from Yelp, reviews with one or two stars are considered negative, reviews with 3 or 4 stars are considered positive.

\section{Implementation}
\paragraph{Training} We use {bert-base-uncased} from huggingface\footnote{\url{https://huggingface.co/transformers/}} for all our experiments. The normal models were fine-tuned for two epochs with a learning rate of 2e-5. We restrict the maximum input length to 128 tokens. For the training with data-augmentation, we train for 12 epochs with a starting learning rate of 2e-5 and linear schedule. We evaluate the robustness on an additional held-out dataset after every epoch. For a threshold of 0.5 on the cosine similarity of words, the robustness reaches its peak after the last epoch. However, we find that two or three epochs are already enough for larger thresholds on cosine similarity of words. All our experiments are conducted on a single RTX 3090.

\paragraph{Attacks} We use TextAttack\footnote{\url{https://textattack.readthedocs.io/en/latest/}} for the implementations of all attacks, including the ones with adjusted thresholds. For adversarial training, we adapt the code from~\citet{meng-geometry}.

\end{document}